\documentclass[letterpaper, 10 pt, conference]{ieeeconf}  
\usepackage{color}

\usepackage{amsmath}
\usepackage{amssymb}
\usepackage{placeins}
\usepackage{cite}
\usepackage{url}
\usepackage{adjustbox}
\usepackage{graphicx}
\usepackage[table]{xcolor}
\usepackage[dvipsnames]{xcolor}
\graphicspath{ {./figures/} }
\renewcommand{\baselinestretch}{0.98}

\IEEEoverridecommandlockouts                              

\title{\LARGE \bf
Robotic Dexterous Manipulation via Anisotropic Friction Modulation using Passive Rollers}

\author{Ethan Fisk$^{1}$, Taeyoon Lee$^{2}$, and Shenli Yuan$^{2*}$
\thanks{$^{1}$Ethan Fisk is with the Department of Mechanical Engineering, Northeastern University, Boston, MA 02115, USA
        {\tt\small fisk.e@northeastern.edu}}%
\thanks{$^{2}$Taeyoon Lee and Shenli Yuan are with the RAI Institute, Cambridge, MA 02142, USA
        {\tt\small \{tlee, syuan\}@rai-inst.com}}%
\thanks{$^{*}$Corresponding author.}%
}

\begin{document}

\maketitle
\thispagestyle{empty}
\pagestyle{empty}

\begin{abstract}
Controlling friction at the fingertip is fundamental to dexterous manipulation, yet remains difficult to realize in robotic hands. We present the design and analysis of a robotic fingertip equipped with passive rollers that can be selectively braked or pivoted to modulate contact friction and constraint directions. When unbraked, the rollers permit unconstrained sliding of the contact point along the rolling direction; when braked, they resist motion like a conventional fingertip. The rollers are mounted on a pivoting mechanism, allowing reorientation of the constraint frame to accommodate different manipulation tasks.  
We develop a constraint-based model of the fingertip integrated into a parallel-jaw gripper and analyze its ability to support diverse manipulation strategies. Experiments show that the proposed design enables a wide range of dexterous actions that are conventionally challenging for robotic grippers, including sliding and pivoting within the grasp, robust adaptation to uncertain contacts, multi-object or multi-part manipulation, and interactions requiring asymmetric friction across fingers. These results demonstrate the versatility of passive roller fingertips as a low-complexity, mechanically efficient approach to friction modulation, advancing the development of more adaptable and robust robotic manipulation.

\end{abstract}
\vspace{-1mm}

\section{Introduction}
\label{sec:intro}

The ability to control friction forces is a key requirement for dexterous manipulation. Human fingers can seamlessly switch between firm grasping and delicate sliding, enabling a wide variety of manipulation actions. Replicating this capability in robotic fingers remains challenging, largely due to the difficulty of accurately modeling contact friction. One strategy to address this challenge is the use of active or passive friction-modulation mechanisms, which are typically designed to provide a clear contrast between high friction for grasping and low friction for sliding.  

Several classes of robotic end-effectors have explored this principle. Active-surface mechanisms use belts, rollers, or conveyors to shift contact locations relative to the finger frame, either through direct actuation of surface motion~\cite{6491289,velvet_finger,ma2016hand,yuan_design_2020-1,yuan_design_2020,cai_-hand_2023,li_active_2024} or by switching contact mechanisms between locked and unlocked states~\cite{8990029,chavan-dafle_two-phase_2015}. Beyond active surfaces, surface friction can also be modulated by alternating between high- and low-friction contact materials~\cite{spiers_variable-friction_2018,lu_origami-inspired_2020}. Another approach is to modulate friction using controlled fingertip vibration~\cite{NAHUM2022105032,10610769,yi_vib2move_2025}, allowing dynamic adjustment of effective friction without changing the underlying contact material.  
Similar goals have been pursued in soft grippers, where friction modulation has been achieved through load-dependent surface deformation~\cite{8722754} or temporary friction reduction via volatile lubricants~\cite{nishimura_soft_2022}. Electrostatic grippers~\cite{fantoni2004design,8202289} represent yet another strategy, leveraging electrostatic forces to generate high stiction and enabling the grasp of objects that are not sensitive to static charge.  Together, these methods highlight the wide range of strategies explored for fingertip friction modulation. 

\begin{figure}[tb!]
\centering
\includegraphics[width = 0.46\textwidth]{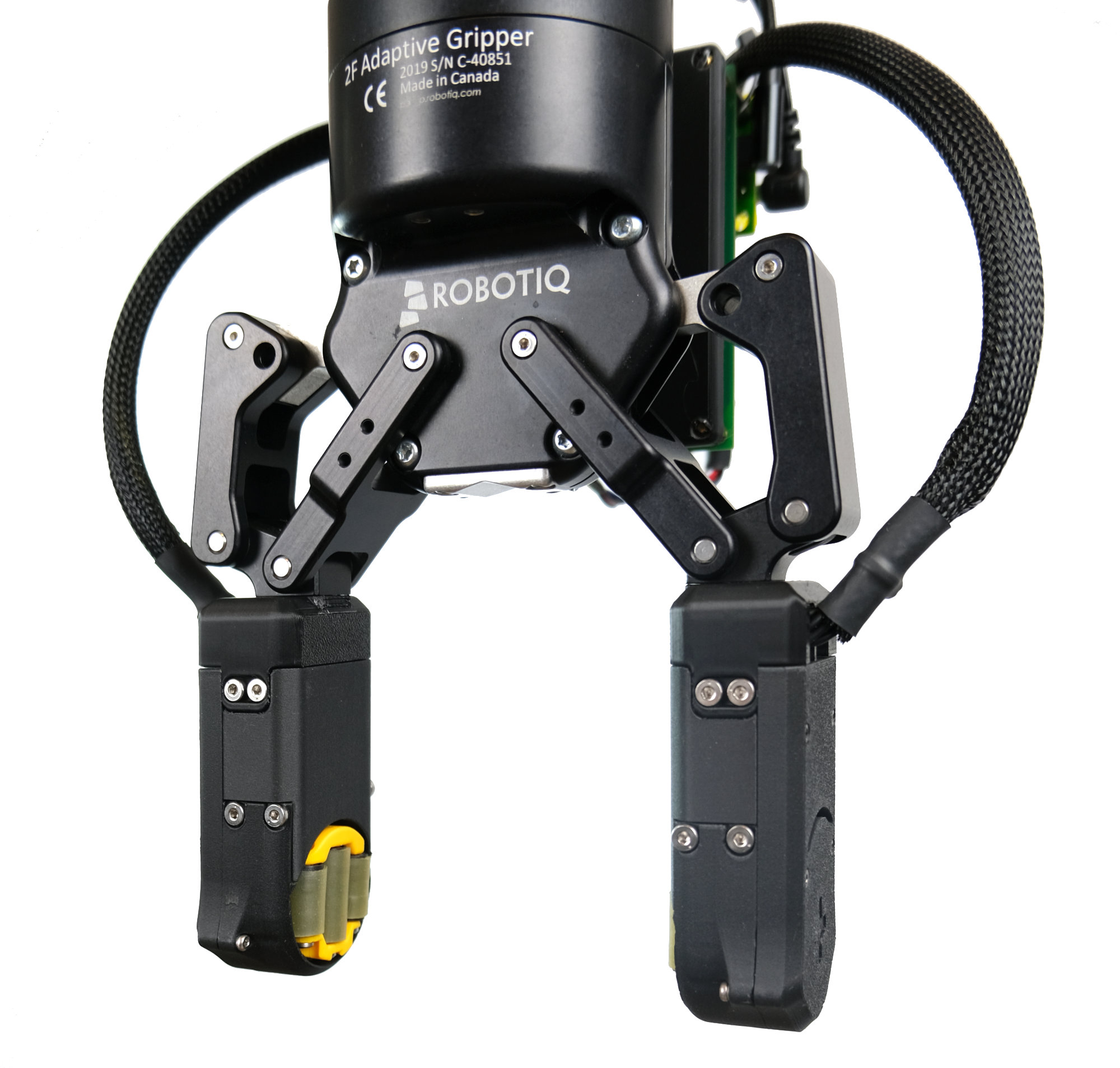}
\vspace{-3mm}
\caption{Prototype fingertips with pivotable, brakeable rollers mounted on a Robotiq 2F-85 Adaptive Gripper.}
\label{fig:cover}
\vspace{-6mm}
\end{figure}

One key benefit of fingertip friction modulation is that it enables manipulation through external contacts and forces. Prior work has extensively explored the use of friction, gravity, and inertial effects beyond the end effector, repeatedly demonstrating that leveraging external forces can yield more robust and dexterous control.  
Research on pushing-based manipulation~\cite{mason_manipulator_1982,lynch_stable_1996,dogar_push-grasping_2010,dogar2011framework,doi:10.1177/0278364919872532,chavan-dafle_stable_2018}, manipulation using gravity~\cite{erdmann_exploration_1988}, controlled slip~\cite{brock_enhancing_1988,karayiannidis2015hand,stepputtis_extrinsic_2018}, and controlled sliding~\cite{kao1992quasistatic,sun1995coordination,howe1996practical,chavan2020planar} can be unified under the concept of \emph{extrinsic dexterity}~\cite{dafle_extrinsic_2014}, demonstrating how external surfaces, gravity, and slip maneuvers can significantly expand the repertoire of robotic hands.
Additionally, in-hand manipulation can also benefit from friction modulation at the contact locations, as demonstrated in~\cite{spiers_variable-friction_2018,teeple_controlling_2022}.


\begin{figure*}[tb!]
\centering
\includegraphics[width=0.85\textwidth]{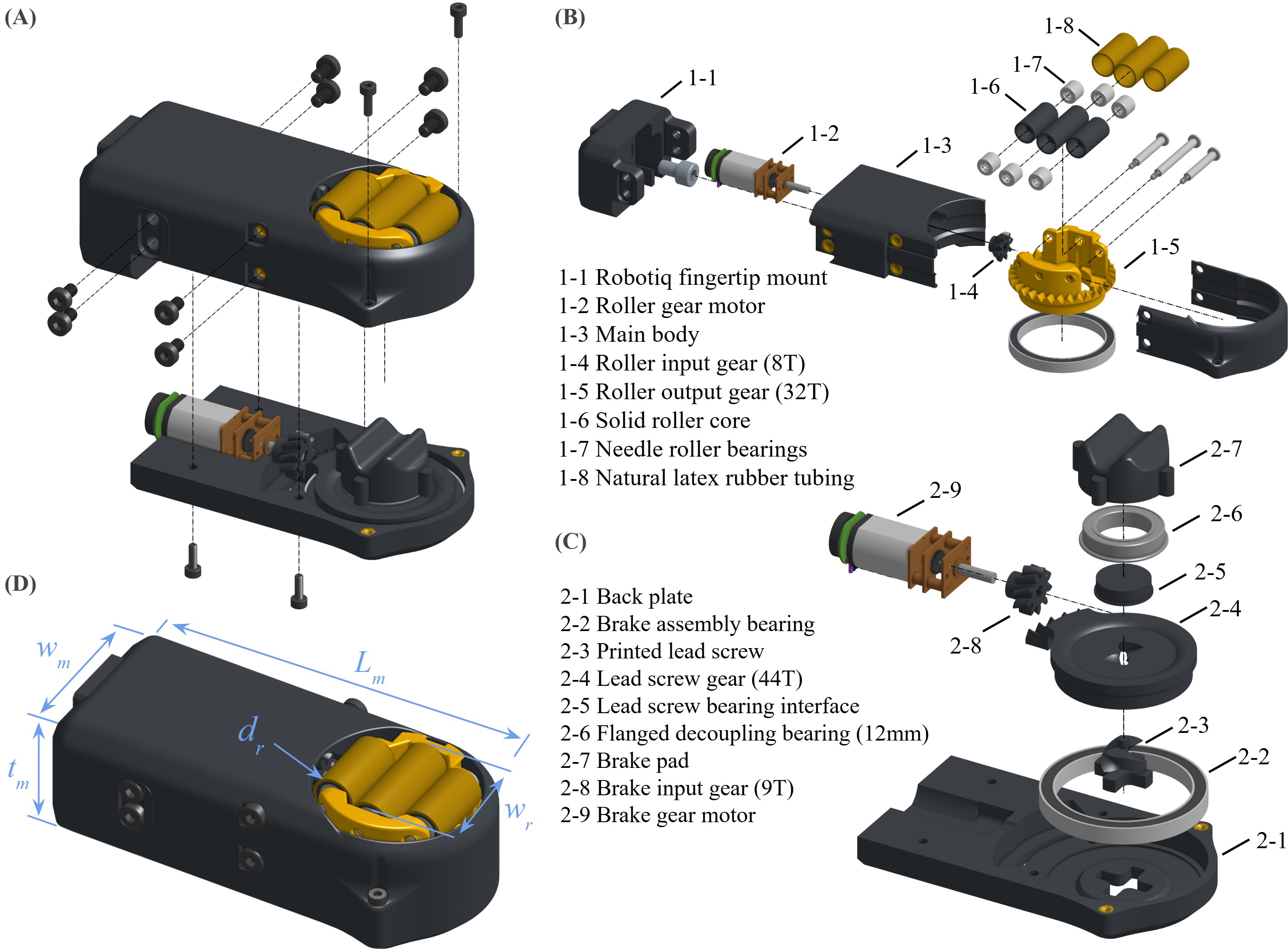}
\vspace{-3mm}
\caption{A CAD model of the proposed roller module: (a) An exploded view showing the separation between brake and roller. (b) An exploded view of the rollers. (c) An exploded view of the brake mechanism. (d) A complete view with call-outs for relevant dimensions.}
\label{fig:design_2}
\vspace{-5mm}
\end{figure*}

In this work, we introduce a fingertip with articulated (reorientable) passive rollers that can be selectively braked or unbraked (as shown in Fig.~\ref{fig:cover}). 
In the unbraked state, the rollers spin freely, enabling unconstrained sliding of the object along the rolling axis.
In the braked state, the rollers function as high-friction surfaces, behaving like soft contacts that resist both linear and torsional motions. Crucially, the rollers can be reoriented independently of the braking mechanism, enabling the constrained direction at the contact to be aligned with the requirements of the manipulation task.

The key contributions of this work are: 
(1) an exploration of how friction modulation can be leveraged for dexterous manipulation;
(2) the design of a novel, mechanically simple fingertip that enables anisotropic friction modulation; and
(3) a kinematic analysis that provides guidance on how manipulation can be performed with the proposed fingertip under environmental constraints.
We also emphasize that this design is not a simplified version of the active roller fingertips~\cite{yuan_design_2020-1,yuan_design_2020}. Instead, it represents a fundamentally different approach to manipulation, focusing on leveraging friction modulation together with environmental constraints.
The remainder of this paper is organized as follows: Section~\ref{sec:design} describes the fingertip design. Section~\ref{sec:model} presents a kinematic analysis of the fingertip integrated with a parallel-jaw gripper. Section~\ref{sec:exp} reports experimental demonstrations of grasping and manipulation tasks, followed by analysis of the results. Section~\ref{sec:conclusion} concludes the paper and outlines directions for future work.

\section{Design}
\label{sec:design}

The proposed roller modules are designed as interchangeable fingertip units for parallel-jaw grippers. Each module consists of three passive rollers, one actuated rotary degree of freedom (DoF) that sets the roller orientation, and an actuated brake that constrains the passive rolling motion when engaged. Both active DoFs are actuated by micro DC motors through $90^\circ$ crown-gear transmissions.

For evaluation, two modules were installed on a Robotiq 2F-85 gripper, selected for its wide adoption and easily swappable fingers. With minor modifications, the design can be adapted for other commercial grippers such as the Franka Hand.


\begin{table}[tb!]
\setlength{\tabcolsep}{10pt}
\renewcommand{\arraystretch}{1.2}
    \caption{Roller Module Dimensions}
    \label{table:roller_dimensions}
    \vspace{-3mm}
    \begin{center}
    \begin{tabular}{|c |c c c c c|}
        \hline
        Property & $w_{m}$ & $t_{m}$ & $l_{m}$ & $d_{r}$ & $w_{r}$ \\
        \hline
        Dimension [mm] & 41 & 27.5 & 92 & 9.2 & 22.5  \\
        \hline
    \end{tabular}
    \end{center}
    \vspace{-7mm}
\end{table}

\subsection{Rollers}
The fingertip module employs reconfigurable roller pads to realize anisotropic friction at contact. A circular contact profile was selected to maintain a consistent contact location during roller reorientation. This profile is approximated with three rollers, one longer central roller flanked by two shorter outer rollers. High anisotropy is achieved by combining a high-friction outer sheath with low-friction rolling elements. Each roller comprises a 3D-printed polymer core, two drawn-cup needle bearings, and a natural-latex tube as the outer layer (see Fig.~\ref{fig:design_2}(B)). The three rollers are mounted on a rotating frame that incorporates an external crown-gear profile and is supported by a captive bearing.

Locating the crown-gear teeth on the exterior of the rotating frame frees interior volume behind the rollers for a brake that must function independently of roller orientation. This layout also reduces overall module width, as the drive motor for roller-frame rotation can be oriented at $90^\circ$ to the frame. The motors used in the module are N20 gear motors with a reduction ratio of $298:1$ and a maximum torque of $\approx 0.5$ Nm. The crown gears provide an additional $4:1$ reduction, resulting in an output torque of $\approx 2$ Nm per finger. 
 
Grouping multiple rollers increases the effective contact area and thereby the resistance to motion in the non-rolling directions, enhancing both lateral and torsional friction. Additionally, employing multiple rollers per fingertip suppresses rotation outside of the roller plane, thereby maintaining the contact normal approximately aligned with the roller plane normal. This configuration mitigates lateral components in the contact reaction force that can arise from local object geometry and reduces the likelihood of grasp failure. This effect is especially critical for passive rollers, as in the unbraked state they exhibit low tangential friction, which would otherwise exacerbate lateral slip and increase the risk of dropping the object.

\subsection{Brake Mechanism}
To modulate fingertip friction, as well as stabilize grasp during fingertip reorientation, a compact brake acts on the backsides of all three rollers (see Fig.~\ref{fig:design_2}(C)). A dedicated gearmotor drives a 3D-printed sector gear whose hub contains an ACME nut profile. Rotation of this sector converts to linear motion via a double-start lead screw ($8$ mm pitch). An anti-rotation cruciform key on the screw end mates with a corresponding feature in the back plate, preventing screw spin. This transmission delivers approximately $1.5$ mm of brake-pad travel for a $35^\circ$ sector rotation, providing a large mechanical advantage between motor rotation and pad displacement and enabling substantial braking force without increasing module thickness. Owing to the gearmotor’s non-backdrivability, the applied clamping force can be maintained without additional holding power. The geared motor, with the $4.88:1$ reduction of the brake mechanism, allows the brake to maintain a theoretical maximum braking force of over $50$ N in an unpowered state. 





The brake pad incorporates two wedge features that engage the gaps between the central and outer rollers, effectively constraining rolling when actuated. To keep the wedges aligned with these gaps as the roller frame rotates, the pad is mounted on a bearing that co-rotates with the frame. Synchronization is achieved with four rounded followers on the pad that loosely engage mating cutouts in the frame, providing rotational coupling while permitting the required axial translation for pad engagement.

\begin{figure}[tb!]
\centering
\includegraphics[width=0.28\textwidth]{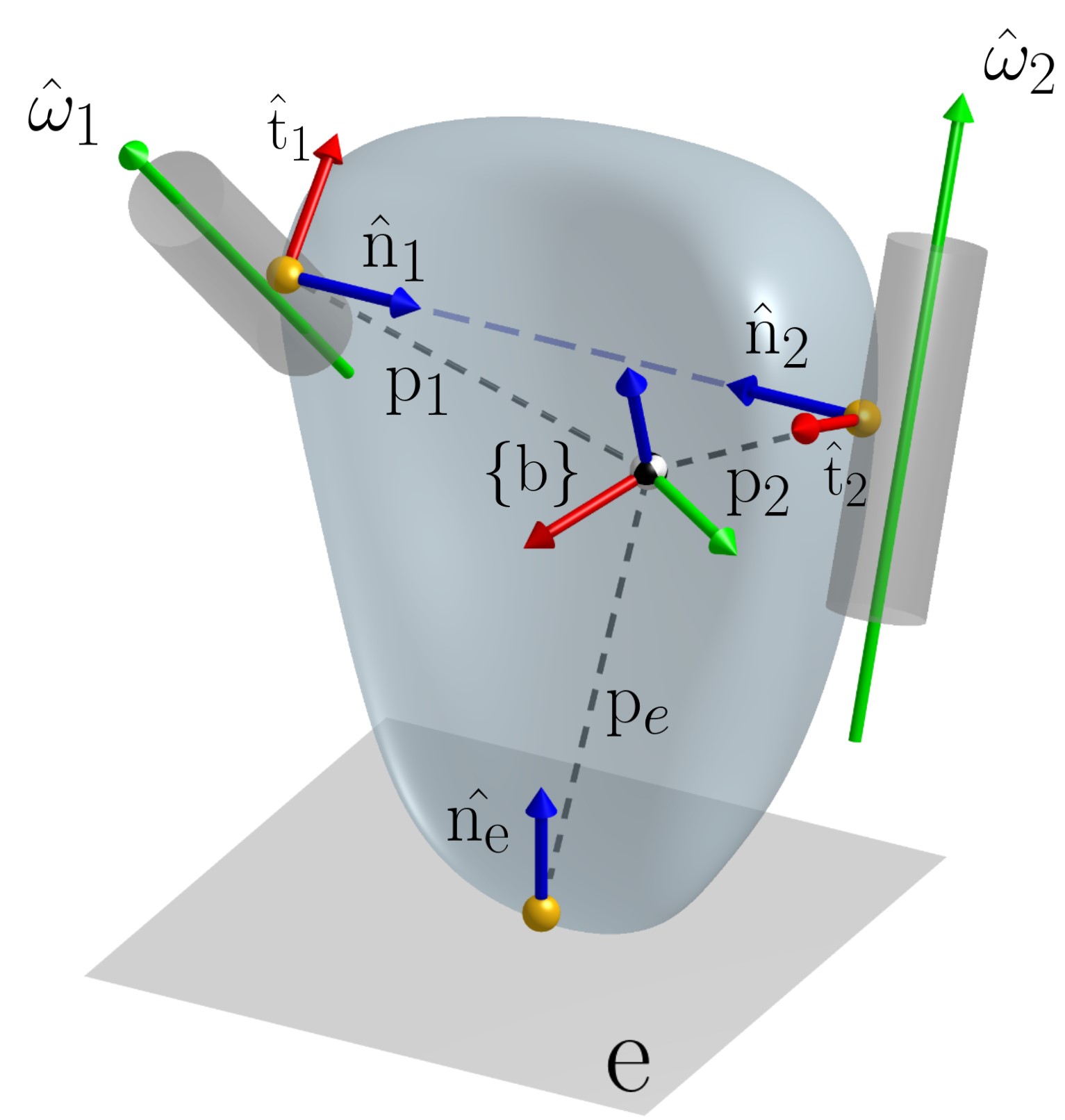}
\vspace{-3mm}
\caption{Kinematic model of a general object under antipodal grasp by the proposed gripper with two arbitrary rolling axes, including environmental contact.}
\label{fig:model}
\vspace{-5mm}
\end{figure}

\section{Kinematic Analysis}
\label{sec:model}
To analyze how the fingertip interacts with a grasped object during manipulation, we conducted a kinematic study using two modules mounted on a parallel-jaw gripper in an antipodal grasp. An object grasped by the proposed setup naturally retains up to two passive DoFs relative to the gripper coordinate frame. This property enables the object to be manipulated in specific ways when additional contact constraints are imposed by the environment. We first present a first-order kinematic analysis of a general object grasped by the gripper, and then extend the discussion to incorporate environmental constraints and end-effector motion, showcasing several classes of object motion realizable by the proposed system.

As shown in Fig.~\ref{fig:model}, the kinematics of an object grasped by two fingertip modules can be modeled as two antipodal rolling contacts, characterized by rolling axes $\hat{\omega}_1, \hat{\omega}_2 \in \mathbb{S}^2$, contact point locations $p_1, p_2 \in \mathbb{R}^3$, and contact normals $\hat{n}_1, \hat{n}_2 \in \mathbb{S}^2$, all expressed with respect to a frame $\{b\}$ attached at the end effector. From the antipodal grasp configuration, it follows that 
\begin{equation}
p_1 - p_2 \perp \hat{\omega}_i ~\text{and} ~ p_1 - p_2 \parallel \hat{n}_i\label{e:antipodal_1}
\end{equation}
for $i=1,2,$ and $\hat{n}_1 = -\hat{n}_2$.

Let $w\in\mathbb{R}^3$ and $v \in \mathbb{R}^3$ denote the instantaneous angular and translational velocities of the object, respectively, at the origin of frame $\{b\}$. Each rolling contact imposes that the velocity of the object at the contact point, given by $v - p_i \times w$ for $i=1,2$, must match the velocity of the roller at that point  \cite{lynch2017modern}. Since a freely spinning roller can only generate surface velocity along the tangential direction $\hat{t}_i := \hat{\omega}_i \times \hat{n}_i$, which is perpendicular to both the rolling axis $\hat{\omega}_i$ and the contact normal $\hat{n}_i$, the following no-slip constraints hold for $i=1,2$:
\begin{align}
    (v - p_i \times w)\cdot \hat{\omega}_i &= 0, \\
    (v - p_i \times w)\cdot \hat{n}_i &= 0.
\end{align}
The soft contact model further imposes, $w \cdot \hat{n}_i = 0$ \cite{lynch2017modern}. Although these relations yield six scalar constraints on the object velocity $V=(w,v)$, only four of them are linearly independent due to the antipodal geometry described in~\eqref{e:antipodal_1}. The resulting four independent constraints on the object velocity can be expressed compactly in matrix form as
\begin{equation}
\begin{bmatrix}
(p_1\times \hat{\omega}_1)^{\top} & \hat{\omega}_1^{\top}  \\
(p_2\times \hat{\omega}_2)^{\top} & \hat{\omega}_2^{\top}  \\
(p\times \hat{n})^{\top} &\hat{n}^{\top} \\
\hat{n}^{\top} & 0 ~~ 0 ~~ 0
\end{bmatrix}
\begin{bmatrix} w \\ v \end{bmatrix} = \begin{bmatrix}
    0\\0\\0\\0
\end{bmatrix}, \label{e:4_constr}
\end{equation}
where $p\in\mathbb{R}^3$ is any point on a line passing through $p_1$ and $p_2$, and $\hat{n} \triangleq \hat{n}_1 = -\hat{n}_2$.

Interestingly, for non-parallel configuration of rolling axes, two canonical basis vectors of the solution space of \eqref{e:4_constr} can be derived, each corresponding to a screw motion whose screw axis passes through the midpoint $q = (p_1+p_2)/2$ of the two contact points. These basis vectors are given by
\begin{align}
    &(\hat{\omega}_{ib},~ -\hat{\omega}_{ib} \times q + h_{ib}\cdot \hat{\omega}_{ib}) , ~ {\text{and}} \label{e:basis_1} \\
    &(\hat{\omega}_{eb},~ -\hat{\omega}_{eb} \times q + h_{eb} \cdot \hat{\omega}_{eb}), \label{e:basis_2}
\end{align}
where $\hat{\omega}_{ib},~\hat{\omega}_{eb} \in \mathbb{S}^2$ denote the internal and external bisectors of the roller axes $\hat{\omega}_1,~\hat{\omega}_2$, and $h_{ib}, h_{eb}\in\mathbb{R}$ are the corresponding screw pitches. Even in the degenerate case where the two rolling axes are parallel, i.e., $\hat{\omega}_{ib} = \hat{\omega}_1=\hat{\omega}_2$ and $\hat{\omega}_{eb} = 0$, two linearly independent basis screw motions still exist. These are (i) a pure rotation about $\hat{\omega}_{ib}$ at the midpoint $q$,
\begin{equation}
(\hat{\omega}_{ib}, -\hat{\omega}_{ib}\times q), \label{e:basis_rot_parallel}
\end{equation}
and (ii) pure translation along the tangential direction $\hat{t} := \hat{\omega}_{ib} \times \hat{n}$,
\begin{equation}
(0, \hat{t}), \label{e:basis_trans_parallel}
\end{equation}
which is consistent with geometric intuition.

Below are some types and cases of additional constraint that eventually fully define the unique motion of the manipulated object.

\subsection{Object geometry}
A non-flat object generally exhibits the full two degrees of passive motion spanned by \eqref{e:basis_1} and \eqref{e:basis_2}; in the special case of a symmetric object such as a sphere, this motion reduces exactly to spherical rotation about the midpoint $q$ while maintaining fixed contact points. In contrast, objects with locally flat contact geometry (e.g., a box) can be fully constrained by the proposed gripper due to higher-order (second-order) contact kinematics. More interestingly, objects with mixed curvature---such as a cylinder, which has one vanishing principal curvature and one nonzero principal curvature---can exhibit a single degree of freedom screw motion corresponding to either \eqref{e:basis_1} or \eqref{e:basis_2}, depending on whether the curved direction aligns with one of the bisectors of the roller axes. Even a locally flat box can retain mobility when the two rollers are parallel, in which case the motion reduces to pure translation along the tangential direction.

It should be noted, however, that we have not yet discussed how the remaining passive degrees of freedom of the object can be actively driven to realize desired motions. In the following section, we describe how additional environmental contacts, combined with nonzero end-effector motion, can determine the object’s motion.

\subsection{Environmental contact and end-effector motion}
Now, suppose the object is in contact with a static environmental surface at point $p_e \in \mathbb{R}^3$ with contact normal $\hat{n}_e \in \mathbb{S}^2$, and that the end effector is driven toward the surface with velocity $V_{ee} = (w_{ee}, v_{ee})$. The relative velocity of the object with respect to the end effector at the contact point $p_e$ must exactly oppose the instantaneous velocity of the end effector at that location in the direction of the contact normal. This can be expressed as the scalar constraint, $(v - p_e \times w) \cdot \hat{n}_e = -\nu_{ee}$, or in matrix form as
\begin{equation}
    \begin{bmatrix}
    (p_e \times \hat{n}_e)^{\top} & \hat{n}_e^{\top}
    \end{bmatrix}
    \begin{bmatrix}
        w \\ v
    \end{bmatrix} = -\nu_{ee}, \label{e:env_contact}
\end{equation}
where $\nu_{ee} \triangleq (v_{ee} - p_e \times w_{ee})\cdot \hat{n}_e$. Note that, unlike the purely passive kinematic constraints in \eqref{e:4_constr}, the relation in \eqref{e:env_contact} necessarily induces nonzero object motion. Moreover, depending on the geometry of the environmental surface---as well as the intrinsic geometry of the grasped object discussed earlier---the object may be subject to additional constraints. To capture these effects in detail, Section~\ref{sec:exp} presents case studies spanning different object geometries, environmental contact conditions, and roller-axis configurations.
 
\subsection{Articulated/complex objects}
Lastly, we observe that the proposed gripper system can manipulate objects with additional internal degrees of freedom in a particularly interesting way. From a purely mobility-oriented perspective, this often requires imposing extra constraints, for example by locking one of the passive rollers to compensate for the additional degrees of freedom introduced by the object’s internal state. From a more pragmatic standpoint, the strategy relies on exploiting asymmetric contact constraints at the two fingers, allowing different parts of the object to be manipulated in distinct ways. Without resorting to a full general kinematic analysis, we illustrate these ideas through case studies on manipulation of the internal state of complex objects in Section~\ref{subsec:asymm_fric}.

\begin{figure}[b!]
\vspace{-5mm}
\centering
\includegraphics[width=0.4\textwidth]{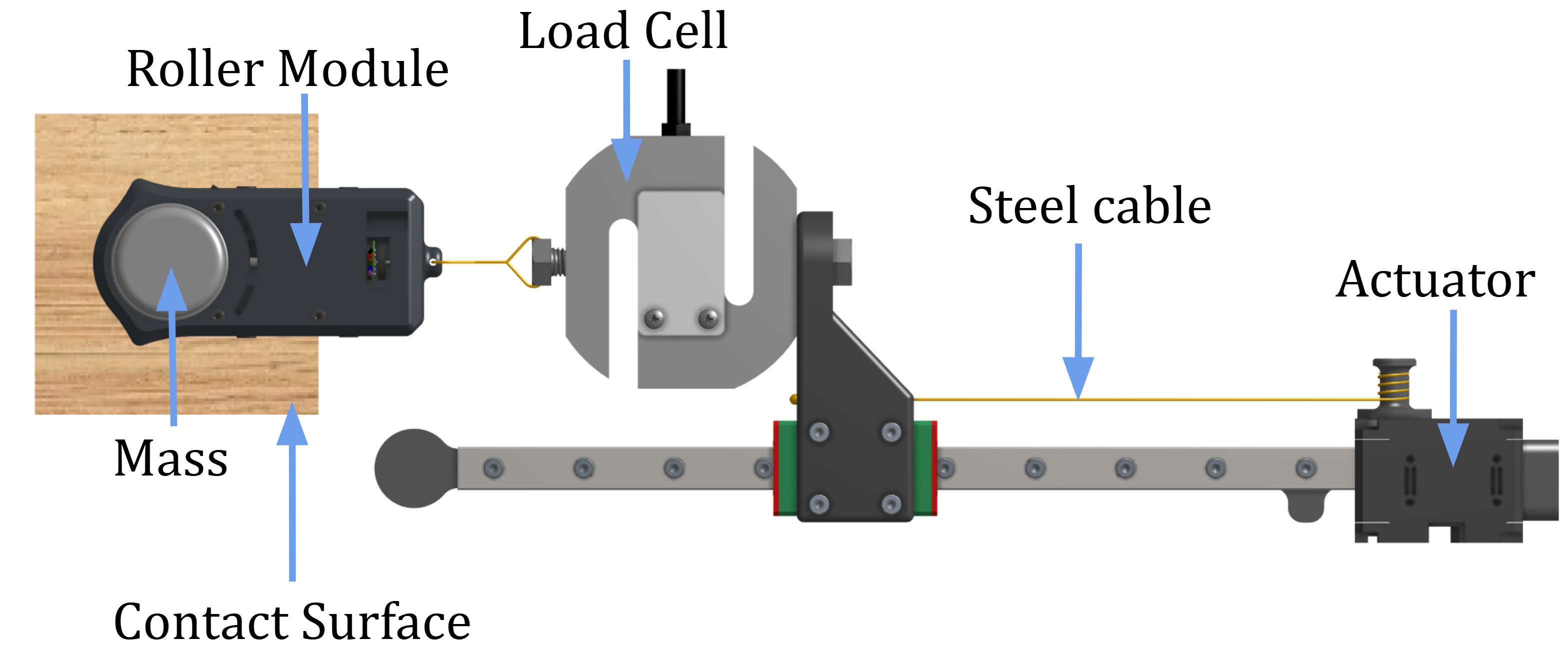}
\vspace{-3mm}
\caption{Friction Test Apparatus (Top View)}
\label{fig:friction_tester}
\end{figure}

\section{Experiments and Discussions}
\label{sec:exp}

\subsection{Friction Measurement}
\label{sebsec:friction}

In order to provide an estimate of the friction variation achievable with the proposed gripper, we conducted experiments to characterize the friction coefficients in both the braked and unbraked states. The breakaway friction coefficients for four different materials were measured to evaluate the rolling friction of the module, and these were measured using a sliding stage (Fig.~\ref{fig:friction_tester}) driven by a Dynamixel actuator (XL430-W250-T). A $1$ kg load cell is mounted to the carriage of the slider and attached at the other end to the roller module using a compliant steel cable. The apparatus was clamped to the test material, and a known weight was placed on the back of the module, with the cable ensuring that this load is transferred exclusively to the rollers. The actuator is then commanded to deliver an increasing load to the slider until slip occurs, at which point the breakaway force is recorded. This was repeated 5 times for each material combination and the rollers were perturbed randomly between each trial. All tests were performed with a $900$ g load and the results of this testing are summarized in Table~\ref{table:friction}. 

\begin{table}[tb!]
\setlength{\tabcolsep}{10pt}
\renewcommand{\arraystretch}{1.2}
\centering
\caption{Breakaway Friction Coefficients}
\label{table:friction}
\vspace{-2mm}
\begin{tabular}{|c|c|c|}
\hline
Material & Rollers braked & Rollers Unbraked \\ \hline 
Plastic & 0.809 ± 0.130 & 0.029 ± 0.015 \\ \hline
Glass & 0.736 ± 0.162 & 0.034 ± 0.009 \\ \hline
Metal & 0.491 ± 0.130 & 0.026 ± 0.013 \\ \hline
Wood & 0.491 ± 0.037 & 0.024 ± .0030 \\ \hline
\end{tabular}
\vspace{-6mm}
\end{table}

The plastic tested was a glossy tabletop, the glass a smooth sheet, the metal an aluminum alloy with a satin finish, and the wood a hardwood with a slightly rough surface. The highest frictions were experienced when in contact with the smooth, high gloss plastic and glass, while the metal and wood surfaces exhibited similarly lower friction coefficients. This outcome is expected, as glossy surfaces provide larger contact areas and distribute pressure more evenly across the rollers. While the braked friction coefficients depend significantly on the surface texture in addition to the material, the unbraked friction coefficients were statistically equivalent across all surfaces, suggesting that the unbraked friction is largely independent of material. Assuming rolling without slipping, the unbraked friction is due to the bearing friction within the rollers, which is not expected to be affected by the material.
In practice, the fingertip torsional friction was observed to be the limiting factor in manipulation through roller pivoting (Section~\ref{subsec:roller_pivoting}). 

It should be noted that both braked and unbraked measurements exhibited noticeable noise, indicating that the friction behavior is not perfectly consistent. One likely cause is roller circularity, as the natural-latex tubing used for the outer layer does not guarantee uniform thickness. Stretching the tubing over solid cores can also introduce high spots, leading to deviations from circularity. This effect is most pronounced in low-force measurements in the unbraked state, where stick–slip behavior represents a significant fraction of the total force. Variability is further compounded by the three-roller configuration, since each roller may differ slightly in geometry and orientation. Over molding or multi-material 3D printing may be a preferable alternative to produce rollers with more consistent properties. 
Despite the variability, the results confirm that the proposed module achieves significant frictional contrast, as the rolling friction coefficients were an order of magnitude lower than the braked coefficients.

\subsection{Manipulation with a Unilateral External Constraint}

As described in Section~\ref{sec:model}, when the rollers are unbraked, the grasped object retains two DoFs. Additional constraints can be imposed through the object’s contact geometry as well as external contacts. 

One of the simplest ways to achieve deterministic manipulation is by introducing an external constraint, such as placing the object on a supporting surface. The surface provides a unilateral one-DoF constraint, while the object’s geometry contributes another. Combined with the four constraints imposed by the rollers, these are sufficient to fully define the resulting object motion.

From here, all the experiments are conducted in a fixed orientation top-down grasp configuration, i.e., $w_{ee} =0$, where the environment contact surface normal $\hat{n}_e$ lines up with the roller axes when their pivot angles $\theta_1, \theta_2$ are zeroed. More specifically, we describe the general orientation of rolling axis as,  $\hat{\omega}_i = \mathrm{Rot}(\hat{n}, \theta_i) \cdot \hat{n}_e$.

\begin{figure}[t]
\centering
\includegraphics[width=0.48\textwidth]{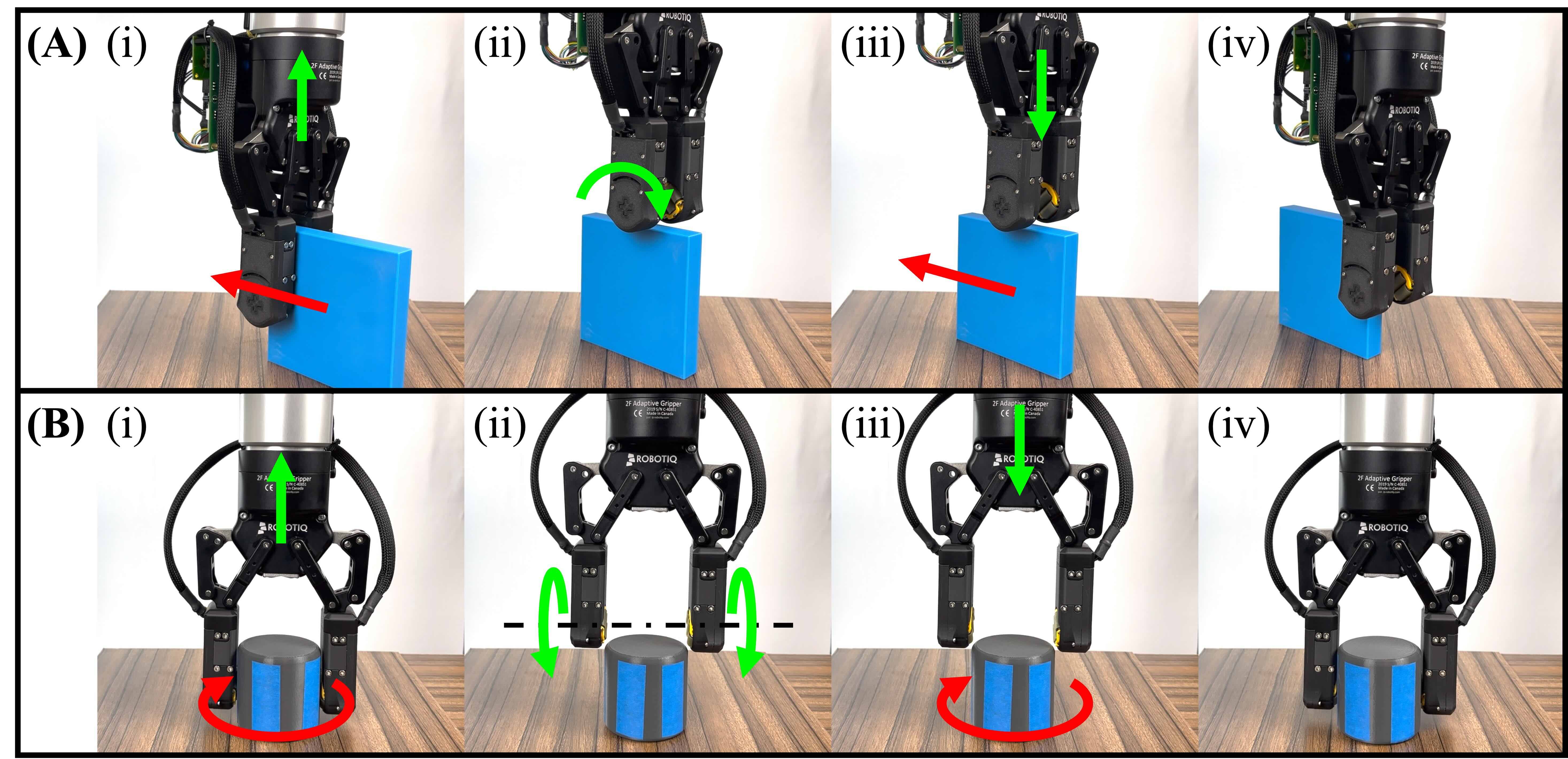}
\vspace{-3mm}
\caption{
\textbf{Object Manipulation with a Unilateral External Constraint.}
Red arrows indicate gripper actions, while green arrows indicate object motion.
\textbf{(A)} Planar Object Translation. 
\textbf{(B)} Cylindrical Object Rotation. 
}
\label{fig:manipulation_slide_screw}
\vspace{-5mm}
\end{figure}

\subsubsection{Object Translation}

Figure~\ref{fig:manipulation_slide_screw}(A) demonstrates a planar object grasped in parallel configuration of roller axes (i.e., $\hat{\omega}_{ib} = \hat{\omega}_1 = \hat{\omega}_2$ and $\theta:=\theta_1 = \theta_2 $) while in contact with a desk. Without environmental contact, the object is constrained to only freely slide along the direction $\hat{t}= \hat{\omega}_{ib} \times \hat{n}$, corresponding to case~\eqref{e:basis_trans_parallel}, as the flat contact geometry prevents rotation about $\hat{\omega}_{ib}$, cf.~\eqref{e:basis_rot_parallel}. Substituting $w = 0$ and $v = \hat{t} \cdot \nu_{obj}$ (where $\nu_{obj}$ denotes the scalar component of object velocity along $\hat{t}$ in the end-effector frame) into the environmental contact constraint~\eqref{e:env_contact} yields 
\begin{equation*}
\nu_{obj} = - \nu_{ee} / (\hat{n}_e^{\top} \hat{t}) = -\nu_{ee} / \sin \theta.
\end{equation*}
When the end-effector only has velocity component parallel to surface normal, i.e., $w_{ee} = 0$ and $v_{ee} = \nu_{ee} \cdot \hat{n}_{ee}$, the resulting object velocity in the global stationary frame, $v_{s} = \nu_{obj}\cdot\hat{t} + \nu_{ee}\cdot\hat{n}_{ee}$ simplifies to
\begin{equation}
v_s = -\nu_{ee}\cdot \sin \theta\cdot \mathrm{Proj}_{e}(\hat{t}), \label{e:planar_parrallel}
\end{equation}
where $\mathrm{Proj}_{e}(\hat{t})$ denotes a projection of $\hat{t}$ onto the tangent plane of environmental contact $e$. Thus, the roller angle determines the travel pitch, i.e., the ratio between the vertical lift of the gripper and the horizontal displacement of the object. A necessary condition for the object to have non-zero translational velocity is that the rolling direction $\hat{t}$ should not coincide with the surface normal (in which case, $\sin(\theta) = 0$). 


In the example shown, two parallel rollers are oriented at $\theta = 45^\circ$ relative to the surface normal. When the gripper moves upward, the object slides leftward along the desk, as illustrated in Fig.~\ref{fig:manipulation_slide_screw}(A)(i,ii). Furthermore, we can preserve the object’s sliding velocity (as in the previous case) by changing the roller orientation from $\theta = +45^\circ$ to $\theta = -45^\circ$ while reversing the direction of the gripper’s vertical motion. Thus, the object continues to move at the same speed, while the gripper now moves downward, as shown in Fig.~\ref{fig:manipulation_slide_screw}(A)(iii,iv). 
\subsubsection{Object Rotation}

A similar principle is demonstrated in Fig.~\ref{fig:manipulation_slide_screw}(B) with a cylindrical object. Here, the two roller axes are not parallel but are rotated away from the contact normal in opposite directions with equal magnitudes, i.e., $\theta := \theta_1 = -\theta_2$. In this configuration, the rollers constrain the object to screw motions, spanned by \eqref{e:basis_1} and \eqref{e:basis_2}, with respect to the end effector frame. 
Because the object is cylindrical, only the screw motion about the axis $\hat{\omega}_{ib} = \hat{n}_e$ (i.e., \eqref{e:basis_1}) is permitted by the contact geometry. As in the previous case, the supporting environmental contact surface provides an additional constraint that relates object's global motion in accordance with the end-effector motion. When the gripper moves vertically, the object performs a screw motion relative to the gripper, while the desk ensures that rotation occurs about the vertical axis (Fig.~\ref{fig:manipulation_slide_screw}(B)(i,ii)). The resulting object velocity $V_s = (w_s, v_s)$ from the global stationary frame can be similarly derived as, 
\begin{equation*}
w_s = \frac{\nu_{ee}}{r \tan \theta}\cdot  \hat{n}_e,~~~ v_s = 0
\end{equation*}
where $r$ is the radius of the cylinder. Here, the vertical motion of the end effector induces pure rotational motion of the object with a pitch determined by $r$ and $\theta$. When the roller orientations are flipped to the opposite sign, the object maintains the same angular velocity even as the gripper reverses its vertical motion. This behavior is illustrated in Fig.~\ref{fig:manipulation_slide_screw}(B)(iii,iv).

\subsection{Uncertain Contact Adaptation}
\subsubsection{Sliding}
\begin{figure}[t]
\centering
\includegraphics[width=0.46\textwidth]{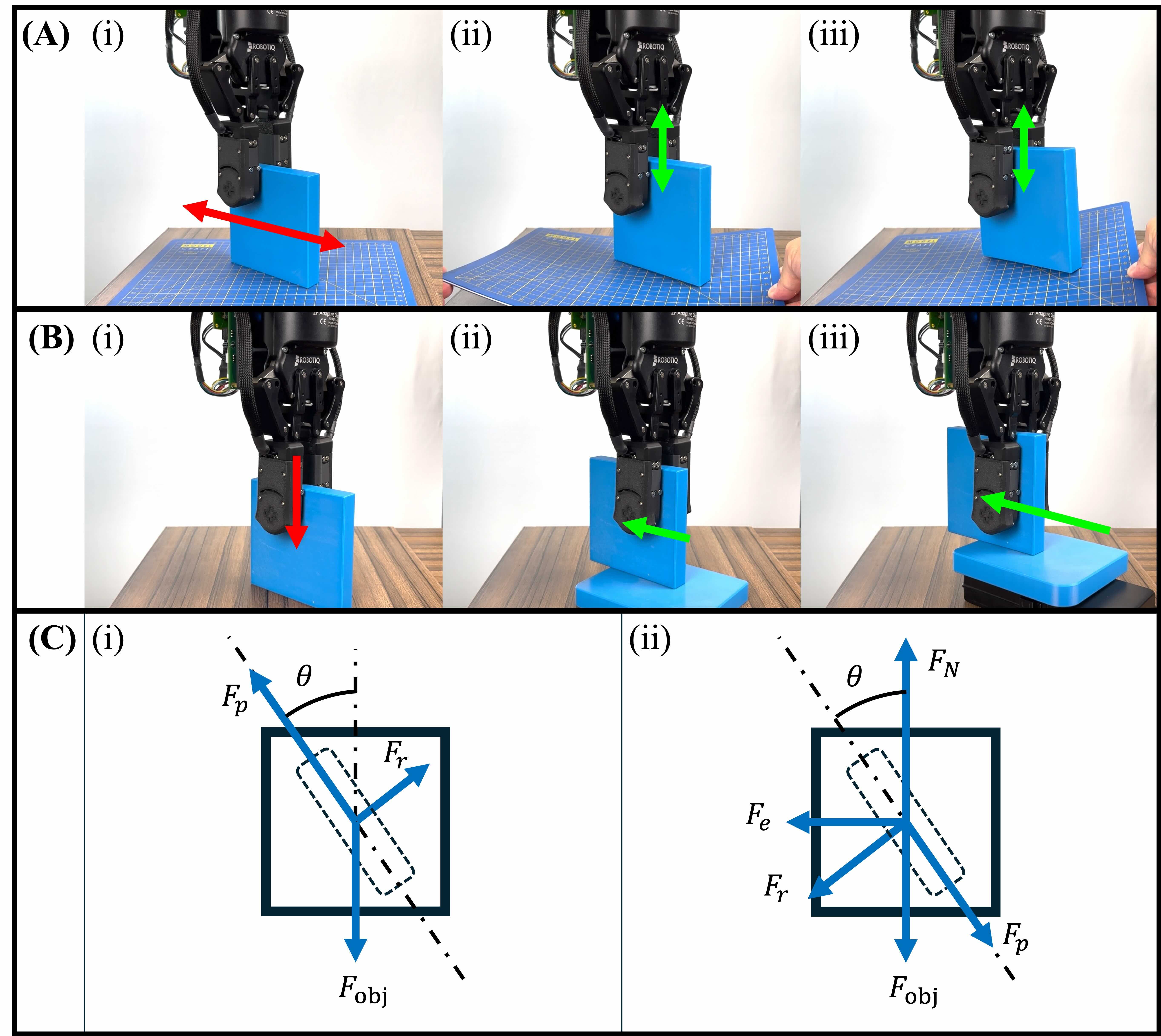}
\vspace{-2mm}
\caption{
\textbf{Object Manipulation with Uncertain Surface Adaptation.}
Red arrows indicate gripper actions, while green arrows indicate the object’s adaptive motion.
\textbf{(A)} Contact Adaptation in Sliding (dragging).
\textbf{(B)} Contact Adaptation in Pick and Place.
\textbf{(C)} Free-body diagram of the object during pick-and-place: (i) in free space, (ii) in contact with the placement surface.
}
\label{fig:manipulation_adaptation}
\vspace{-5mm}
\end{figure}

From~\eqref{e:planar_parrallel}, when $\theta :=\theta_1=\theta_2 \pm \pi/2$, the relationship between the vertical motion of the end effector $\nu_{ee}$ and the horizontal motion of the object becomes decoupled, since $\mathrm{Proj}_e(\hat{t}) = 0$. In this case, the horizontal motion of the end-effector translates directly to that of the object, while vertical motion of the object remains entirely passive.
In other words, when the rollers are oriented in this way, the object can be dragged along the supporting surface regardless of the gripper’s relative distance from the surface. As a result, manipulation becomes highly robust to variations in the surface profile and can easily adapt to changes in surface altitude. This behavior is demonstrated in Fig.~\ref{fig:manipulation_adaptation}(A). It is worth noting that adaptation occurs whenever the rollers are not oriented vertically (Fig.~\ref{fig:manipulation_slide_screw}). However, when the rollers are angled, the adaptation motion acquires a horizontal component and thus influences the object’s horizontal dragging motion. In contrast, with horizontally oriented rollers, as in this example, motion transfer between the gripper and the object is maximized, since the adaptation direction is decoupled from the dragging direction.

An alternative interpretation is that the gripper–object interaction exhibits low impedance in the direction perpendicular to the roller axes and high impedance in all other directions. Consequently, when the object is perturbed by changes in external contacts along the low-impedance direction, it can adapt with minimal resistance from the gripper.

\subsubsection{Pick-and-Place}



Adaptation can also be achieved when the intended motion aligns with the perturbation direction. For example, in a simple pick-and-place task, the object may move vertically while the uncertainty is also along the vertical axis. If the placement surface is higher than expected, the object makes contact earlier than planned. In such cases, the rollers can be oriented so that the object adapts in a direction not aligned with the perturbation. Here, the adaptation motion and the perturbation magnitude are coupled through the roller orientation.  

Figure~\ref{fig:manipulation_adaptation}(B) illustrates this scenario. When the object is placed on a placement surface that is higher than anticipated, it makes early contact while the gripper continues moving downward. With the rollers properly oriented, the object can slide laterally along the surface, accommodating the gripper’s overtravel. To achieve this behavior reliably, approximate estimates of contact friction and basic force control of the gripper are required.  

Figure~\ref{fig:manipulation_adaptation}(C) provides a static analysis for the object in free space and when in contact with the placement surface. A successful pick-and-place requires the object not to slide along the rolling direction when lifted, as shown in Fig.~\ref{fig:manipulation_adaptation}(C)(i). In this case, we have  
\begin{equation}
\label{equ:pnp_1}
    F_{\text{obj}} \sin \theta = F_r < \mu_r F_g
\end{equation}
where $F_{\text{obj}}$ is the weight of the object, $F_r$ is the roller friction force, $\mu_r$ is the roller’s friction coefficient, and $F_g$ is the gripping force.  This condition establishes an upper bound on $\theta$ for a successful pick:
\begin{equation}
    \sin \theta < \mu_r \frac{F_g}{F_\text{obj}}
\end{equation}

Since the goal is to enable contact adaptation, we must ensure that when the object is pushed against the surface it will slide rather than transmit excessive force. As shown in Fig.~\ref{fig:manipulation_adaptation}(C)(ii), we have  
\begin{align}
\label{equ:pnp_2}
    (F_N - F_{\text{obj}}) \sin \theta &= F_r + F_e \cos \theta \notag \\
    &= \mu_r F_g + \mu_e F_N \cos \theta
\end{align}
where $F_N$ and $F_e$ are the normal and frictional forces applied by the placement surface, respectively, and $\mu_e$ is the surface friction coefficient. 

Assuming the object is placed along the surface normal, $F_e \cos \theta > 0$. Combining this with~\eqref{equ:pnp_1} yields $F_N > 2 F_{\text{obj}}$, meaning the normal force from the surface must be at least twice the object’s weight if sliding occurs during pick-and-place. Notably, this still represents an improvement over rigid gripping, where unexpected contact could transmit significantly larger forces.  

A further relationship can be derived from ~\eqref{equ:pnp_1} and~\eqref{equ:pnp_2}:  
\begin{gather}
\label{equ:pnp_3}
    \tan \theta > \mu_e \left ( 1 - \frac{2F_{\text{obj}}}{F_N} \right)^{-1} > 0.
\end{gather}
This inequality establishes a lower bound on $\theta$. A higher surface friction coefficient $\mu_e$ requires a larger $\theta$ for the condition to be satisfied.

\begin{figure}[t]
\centering
\includegraphics[width=0.47\textwidth]{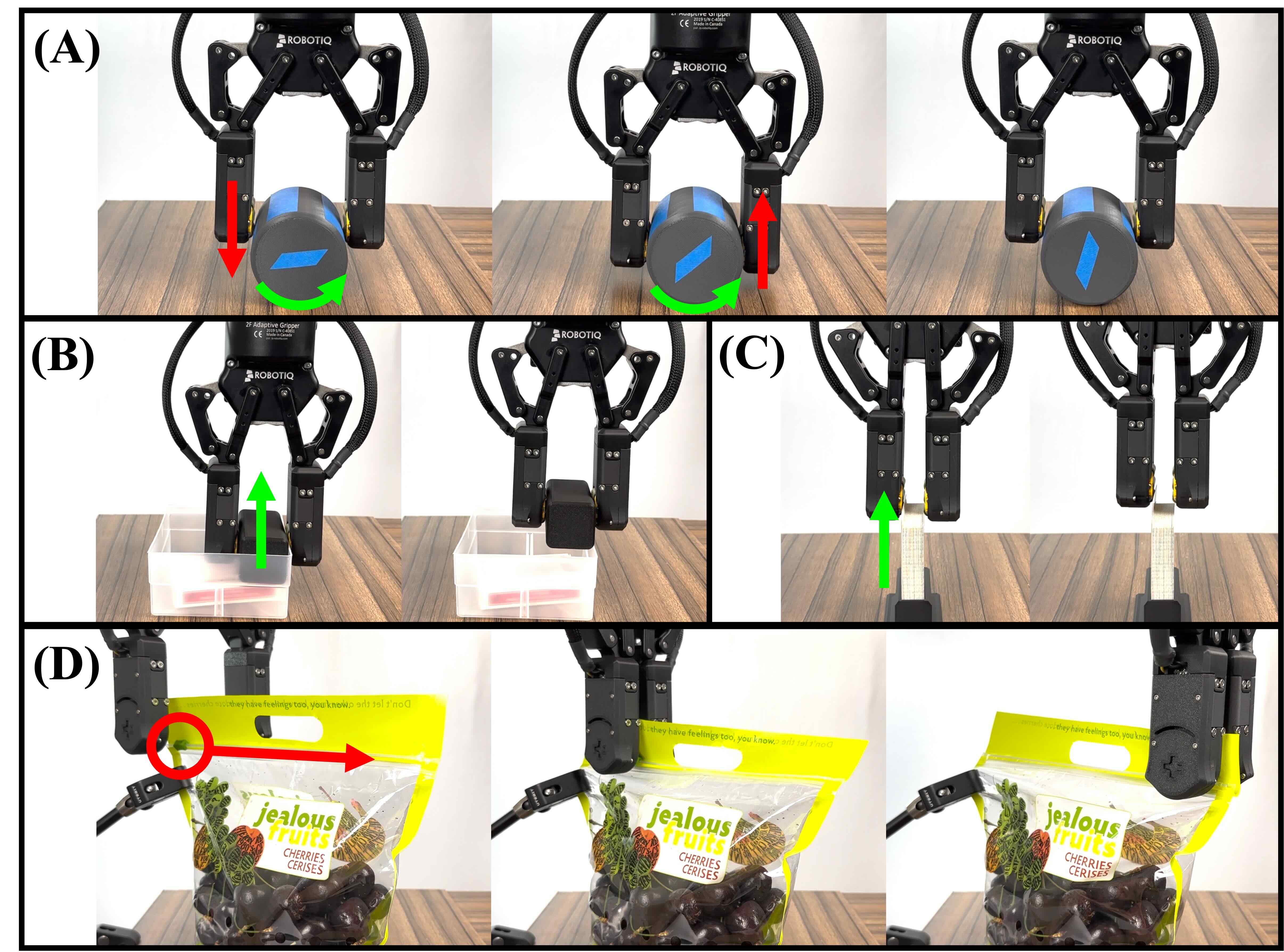}
\vspace{-2mm}
\caption{
\textbf{Manipulation with asymmetric finger friction.}  
\textbf{(A)} Cylinder rolling: left roller braked as the gripper moves downward; right roller braked as the gripper moves upward.  
\textbf{(B)} Object retrieval from a container: left finger (braked) contacts the object, while the right finger (unbraked) contacts the container wall.  
\textbf{(C)} Card extraction from a deck: left finger braked and right finger unbraked, enabling pickup of the leftmost card.  
\textbf{(D)} Closing a zip bag slider: left finger (braked) contacts the slider, while the right finger (unbraked) supports the back of the bag.  
}
\label{fig:manipulation_diff_friction}
\vspace{-5mm}
\end{figure}

\subsection{Manipulation with Asymmetric Finger Friction}
\label{subsec:asymm_fric}
In many practical scenarios, humans naturally apply different levels of friction with different fingers. Typically, the finger(s) providing higher friction drive the object’s motion, while the finger(s) with lower friction act as stabilizers, anchoring the object without resisting its intended movement. This ability to modulate friction across fingers not only enables diverse manipulation strategies but also facilitates the handling of multiple objects or multi-part objects.  

Figure~\ref{fig:manipulation_diff_friction} illustrates several examples of manipulation using asymmetric fingertip friction. In Fig.~\ref{fig:manipulation_diff_friction}(A), a cylinder lying sideways on a surface is rolled by alternating between braked and unbraked rollers. The braked rollers apply high friction, constraining the object with the same linear velocity at the contact point, while the unbraked rollers slip. By alternating which side is braked and coordinating up–down gripper motion, the cylinder can be rolled continuously within the grasp.  

Beyond manipulating a single object with differential friction, similar strategies can be applied when the end-effector interacts with multiple objects. For example, in Fig.~\ref{fig:manipulation_diff_friction}(B), the gripper retrieves an object from a narrow container. A natural human strategy is to slide the object along the container wall while pulling it out—a motion difficult for robots due to the challenge of independently modulating fingertip friction. Here, one finger with a braked roller constrains the object, while the unbraked roller remains outside the container to provide additional support and gripping force as the object is extracted. 
A related strategy is shown in Fig.~\ref{fig:manipulation_diff_friction}(C), where only the card in contact with the braked roller is lifted from a deck, leaving the rest behind. Even soft or flexible objects, such as a zip bag with a slider, can be manipulated in this way (Fig.~\ref{fig:manipulation_diff_friction}(D)): the braked rollers drive the slider while the unbraked rollers slide along the opposite side of the bag to provide support. For floppy objects of this type, the unconstrained fingertip support is critical for the slider to move successfully.

\begin{figure}[t]
\centering
\includegraphics[width=0.46\textwidth]{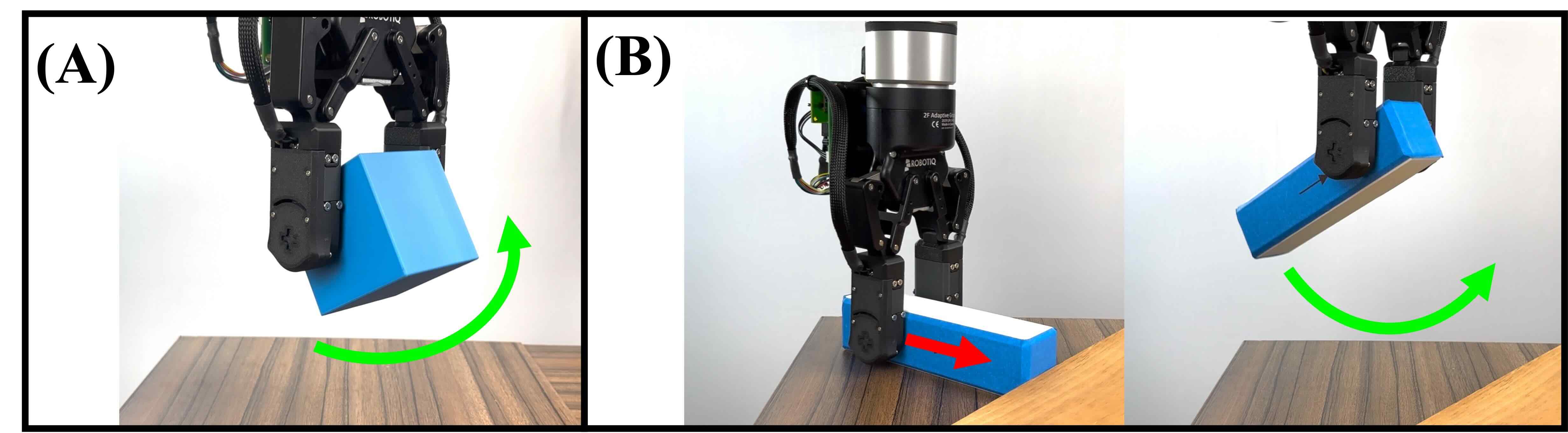}
\vspace{-2mm}
\caption{
\textbf{Manipulation using roller pivoting.} 
\textbf{(A)} Basic pivoting. 
\textbf{(B)} Long-object manipulation combining pivoting with other strategies.
}
\label{fig:manipulation_pivot}
\vspace{-5mm}
\end{figure}

\subsection{Roller Pivoting}
\label{subsec:roller_pivoting}

In most of the aforementioned cases, manipulation is achieved by applying different constraints to the object, with roller pivoting used to set the constrained directions. However, since the rollers themselves provide actuated degrees of freedom at the fingertips, pivoting can also be directly exploited for manipulation by actuating both joint pivots (Fig.~\ref{fig:manipulation_pivot}(A)), following a strategy similar to that presented in~\cite{yuan_design_2020-1}.  
Moreover, roller pivoting can be combined with the other strategies described above to accomplish more complex manipulations, as illustrated in Fig.~\ref{fig:manipulation_pivot}(B). In this case, the object is too long to be continuously pivoted within the grasp. Instead, pivoting is used to adjust the grasp pose, while within-grasp sliding is used to shift the grasp location. By alternating between these two actions, the gripper can achieve continuous manipulation of an oversized object.




\subsection{Limitations}

\subsubsection{Brake Speed}  
The current lead screw brake is limited in speed by the motor and transmission.
Although the brake time is only about $0.7$ seconds, 
a solenoid-based mechanism could provide more instantaneous switching. 

\subsubsection{Contact Area}  
Although the three rollers can form multiple contacts with the object, their small size limits the contact area, potentially reducing grasp stability.

\subsubsection{Object Suitability}
We assume stable object–fingertip contacts; complex object surface geometries may affect the effectiveness of the proposed design.


\section{Conclusions}
\label{sec:conclusion}

This work presented the design and analysis of a robotic fingertip capable of modulating contact friction by braking and unbraking passive rollers embedded at the fingertip. Each fingertip incorporates three passive rollers which, when unbraked, allow the contact point to roll freely over the object surface—effectively enabling unconstrained sliding along the rolling direction. The rollers are mounted on a pivoting mechanism, allowing the fingertip to impose varying constraint directions on the grasped object.  
We provided a constraint-based analysis of the mechanism integrated into a parallel-jaw gripper and demonstrated through experiments that the fingertip can perform a wide range of dexterous manipulations derived from this analysis. The design showed strong versatility, enabling tasks that are conventionally difficult, such as multi-object handling and robust manipulation under uncertainty.  
Future work will focus on improving the mechanism to achieve faster transitions between braking and unbraking, as well as extending the approach to dynamic manipulation under constraints.





\bibliographystyle{IEEEtran}
\bibliography{references}

\end{document}